\documentclass[letterpaper, 10 pt, conference]{ieeeconf} 
\IEEEoverridecommandlockouts                              
\renewcommand{\baselinestretch}{0.95}
\usepackage[T1]{fontenc} 
\usepackage{graphicx} 
\graphicspath{{figures/}} 
\usepackage[bookmarks=true,hidelinks]{hyperref}
\usepackage{amsmath} 
\usepackage{amssymb}  
\usepackage{siunitx}
\usepackage{mathtools}
\usepackage{caption}
\usepackage{subcaption}
\usepackage{booktabs}
\usepackage[utf8]{inputenc}
\usepackage[capitalise]{cleveref}
\usepackage[dvipsnames]{xcolor}
\usepackage{pgfplots}
\usepackage{lipsum} 
\DeclareUnicodeCharacter{2212}{−}
\usepackage[style=ieee, url=false, doi=false, natbib=true, mincitenames=1, maxcitenames=1, maxbibnames=99]{biblatex}
\usepackage{tikz}
\usetikzlibrary{patterns,shapes.arrows, arrows, positioning, decorations.pathreplacing, chains, calc, backgrounds, fit}
\usepackage{stackengine}
\newcommand\solidcirc[4][0]{\rotatebox{#1}{\tikz{\draw[line width=#2] (0,0) 
  arc [x radius=#3,y radius=#4,start angle=0,end angle=360];}}}
\pgfplotsset{compat=newest, every tick label/.append style={font=\footnotesize}}
\usepackage[keeplastbox]{flushend}
\title{\LARGE \bf
Dynamics-Aware Spatiotemporal Occupancy \\ Prediction in Urban Environments
}
\author{Maneekwan Toyungyernsub, Esen Yel, Jiachen Li, and Mykel J. Kochenderfer
\thanks{The authors are with the Stanford Intelligent Systems Laboratory (SISL), Stanford University, CA, USA. 
Email: {\tt\footnotesize \{maneekwt, esenyel, jiachen\_li, mykel\}@stanford.edu}.}
}
\begin{document}
\maketitle
\thispagestyle{empty}
\pagestyle{empty}
\tikzstyle{block}=[align=center,fill=gray!10,rounded corners,draw=black]
\begin{abstract}
Detection and segmentation of moving obstacles, along with prediction of the future occupancy states of the local environment, are essential for autonomous vehicles to proactively make safe and informed decisions.
In this paper, we propose a framework that integrates the two capabilities together using deep neural network architectures. Our method first detects and segments moving objects in the scene, and uses this information to predict the spatiotemporal evolution of the environment around autonomous vehicles. To address the problem of direct integration of both static-dynamic object segmentation and environment prediction models, we propose using occupancy-based environment representations across the whole framework. Our method is validated on the real-world Waymo Open Dataset and demonstrates higher prediction accuracy than baseline methods.   
\end{abstract}

\section{INTRODUCTION}
To achieve safe and reliable autonomous vehicle (AV) navigation, the ability to accurately predict the temporal evolution of the environment surrounding the AVs is important. 
It allows the AVs to intelligently plan safe trajectories based on the other moving obstacles' anticipated behavior.

In this spatiotemporal environment prediction task, the occupancy grid map (OGM)~\cite{Elfes} is commonly used to represent the environment around the AVs. The OGM representation can be obtained from sensor measurements, e.g. on-board LiDAR sensor readings. The OGMs discretize the environment into grid cells and consider the binary free or occupied hypotheses. Each cell in the OGMs contains the belief of its respective occupancy probability.
An alternative representation is evidential occupancy grid map (eOGM)~\cite{Pagac}, where each cell carries an additional information on the occluded occupancy hypothesis in its information channel, in addition to the occupied and free channels. Hence, the OGM representation is akin to RGB images in its discretized spatial structure with information stored in respective channels. 
Due to this similarity, the task of predicting the future states of the environment can be posed as a video frame prediction task and the neural network architectures designed for video frame prediction can be used to predict the temporal evolution of environments represented as OGMs~\cite{Masha, Schreiber, Mohajerin_2019_CVPR, mery_icra, Lange_iros}.

\citet{Masha} repurpose the Predictive Coding Network (PredNet)~\cite{PredNet}, based on a convolutional long-short term memory (ConvLSTM)~\cite{convlstm} architecture, to predict the future environment states. This approach captures the relative motion between static and dynamic objects well, but suffers from vanishing dynamic objects in the predictions at longer time horizons. \citet{mery_icra} develop a double-prong model based on the PredNet model to incorporate environment dynamics within the architecture. One prong receives as input the static OGMs, which contain only the globally static parts of the environment, and the other prong receives the dynamic OGMs, which contain only the moving objects in the environment. The static prong learns to predict the motion model for the globally static parts, i.e. learning the relative motion of the static environment with respect to the moving AV. Likewise, the dynamic prong learns to predict the dynamic parts of the environment. The outputs from these two prongs are fused to produce the complete OGM predictions using Dempster-Shafer Theory (DST)~\cite{dst}.   

\citet{mery_icra} find that the double-prong model is able to accurately predict the motion of dynamic objects at longer prediction time horizons, thereby reducing the issue of predicted object disappearances found in prior work~\cite{Masha}. However, the main limitation of this method is the need for highly accurate object detection and tracking information. Since the inputs to the double-prong model consist of separate static and dynamic OGMs, there is a need to determine which parts of the environment are static and dynamic, respectively, in advance.  

\begin{figure*}[t]
    \centering
    \input{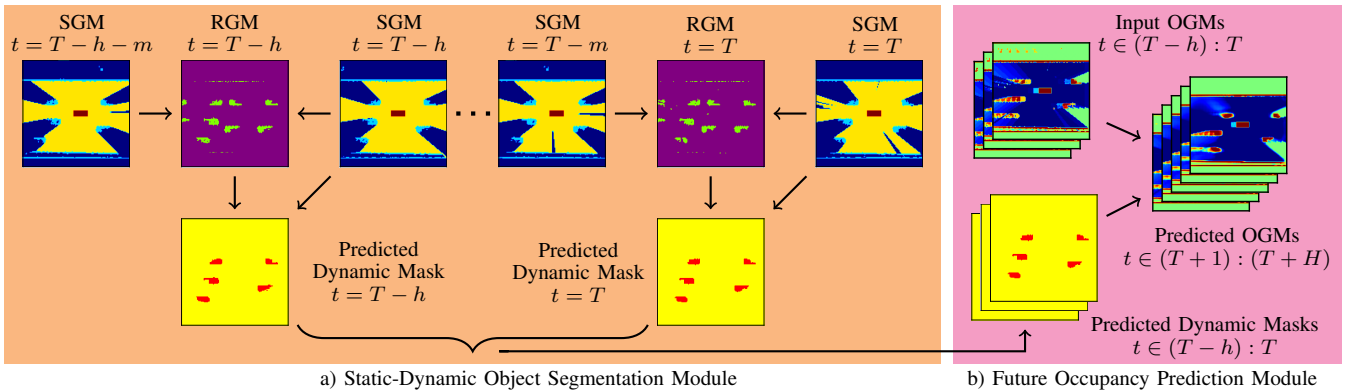}
    \vspace{-1em}
    \caption{\small 
    Static-dynamic object segmentation is performed to distinguish between the static and actually moving objects in the environment. The predicted static-dynamic segmentation outputs are subsequently used as inputs to predict the future occupancy states.}
    \label{fig:teaser}
\vspace{-1.5em}
\end{figure*}

We extend the work by~\citet{mery_icra} by incorporating static-dynamic object segmentation for the spatiotemporal occupancy prediction task. We aim to eliminate the need for object detection and tracking information to be known a priori. Inspired by prior work on moving object segmentation~\cite{mos}, we repurpose the SalsaNext~\cite{salsanext} neural network architecture designed for semantic segmentation to instead detect static and moving objects in the environment. The proposed integration of static-dynamic object segmentation and future occupancy prediction is illustrated in~\cref{fig:teaser}.

Our contributions are as follows. We develop a method that integrates static-dynamic object segmentation and local environment prediction together, without assuming knowledge of static and dynamic objects in the scene. We propose using an occupancy-based environment representation across the entire system to enable direct integration. With our proposed input representations, we are able to accurately segment static and dynamic objects. We validate our method on the real-world Waymo Open Dataset~\cite{Sun_2020_CVPR}, and show that our model outperforms the baselines in predicting the future occupancy states of the environment, without using ground truth static and dynamic object knowledge. 

\section{RELATED WORK}
\label{related work}
\subsection{Occupancy Grid Prediction}
Following successes with the video frame prediction task using deep learning architectures, many have adapted these architectures to predict the future states of the environment~\cite{Masha, Schreiber, Mohajerin_2019_CVPR, 
deep_tracking, mery_icra, Lange_iros}.
\citet{deep_tracking} employ a recurrent neural network (RNN) architecture for occupancy grid prediction. 
\citet{Schreiber} use a ConvLSTM~\cite{convlstm} architecture to predict the future environment represented as dynamic occupancy grid maps (DOGMas)~\cite{Nuss_DOGMa}. The DOGMa representation, however, requires explicit estimation of the cell-wise velocities using a particle filter, which is very computationally expensive. In contrast, our work aims to efficiently classify static and dynamic objects in the scene without resorting to computing cell-wise velocity estimates.  

\citet{Masha} propose a framework using the PredNet~\cite{PredNet} architecture for future environment prediction, but this approach suffers from dynamic object disappearance in the predictions at longer time horizons. 
Recent attempts to prevent vanishing work by alternatively incorporating environment dynamics directly within the architecture~\cite{mery_icra}, or by augmenting attention mechanism to the ConvLSTM architecture~\cite{Lange_iros}. \citet{mery_icra} propose a double-prong model based on the PredNet~\cite{PredNet} architecture, consisting of the static and dynamic prongs. Since the inputs to these two prongs are separate static OGMs and dynamic OGMs, this method requires the static and dynamic objects to be known a priori. This necessitates access to highly accurate object detection and tracking information. Our approach does not require access to such information. Additionally, the moving object identification in~\cite{mery_icra} requires thresholding object displacements between two time steps, which needs to be tuned for various objects to account for different average speeds (e.g. pedestrians and cars).

\subsection{Static-Dynamic Object Segmentation}
With numerous successes on the semantic segmentation task using deep learning, recent work have adapted semantic segmentation architectures for moving object segmentation task. \Citet{Shi_2020_CVPR} develop a 3D point cloud-based semantic segmentation architecture that also predicts dynamic objects by exploiting sequential point clouds. However, an approach based on operating directly on point cloud data is usually very computationally expensive, as well as difficult to train. ~\citet{mos} adapt range projection-based semantic segmentation architectures for dynamic object segmentation. Range images are used as an intermediate representation of the point clouds to reduce computational complexity. They experiment wtih various semantic segmentation architectures and find that the SalsaNext~\cite{salsanext} network performs the best. Additionally, an increase in performance is achieved when providing the network with LiDAR-based residual images, generated by computing the normalized absolute difference between the ranges of the current range image and the range image at previous time steps. Their best performance is realized when training the model with the current range image and multiple residual images. Inspired by~\citet{mos}, we adapt the SalsaNext~\cite{salsanext} architecture with our proposed occupancy-based environment representation in the forms of sensor grid maps (SGMs) and residual grid maps (RGMs), instead of range images, for direct integration with the occupancy grid prediction model as shown in \cref{fig:teaser}.

\section{APPROACH}
\label{approach}
\begin{figure}
    \centering
    \begin{tikzpicture}[>=stealth',font=\footnotesize]
\matrix [column sep=0.16cm,row sep=0.15cm] {
\node[block] (pointcloud) {Point\\Clouds}; &  
\node[block] (envrep) {Occupancy-\\Based\\Representations}; & 
\node[block] (dynobjseg) {Static-Dynamic\\Object\\Segmentation}; &
\node[block] (envpred) {Occupancy\\Prediction}; &
\node[block] (predeogm) {Predicted\\OGMs}; \\
};
\draw [->] (pointcloud) edge (envrep)
    (envrep) edge (dynobjseg)
    (dynobjseg) edge (envpred)
    (envpred) edge (predeogm)
;
\end{tikzpicture}
    \vspace{-1.5em}
    \caption{\small Pipeline for the proposed framework. The dynamic object segmentation module predicts the static and dynamic parts of the environment. The environment prediction module outputs the future OGM predictions.}
    \vspace{-2em}
    \label{fig:pipeline}
\end{figure}
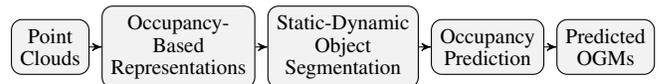

In this study, we aim to develop a dynamics-aware spatiotemporal environment prediction system to predict the future OGMs, without relying on access to object detection and tracking capabilities.
The proposed framework is described in this section, as illustrated in \cref{fig:pipeline}. 
Range-bearing sensor data, e.g. LiDAR, are used to form the environment representation. As done by \citet{Masha}, we first perform ground point removal on point clouds using a Markov Random Field~\cite{Postica} to remove points belonging to the ground. 
We consider splitting the system into two modules.
The static-dynamic object segmentation module determines the parts of the environment that are static and dynamic. 
The downstream environment prediction module then predicts the future OGMs, incorporating the dynamic information obtained from the former module.
In contrast with prior work~\cite{mery_icra}, we incorporate dynamic information without assuming that static and dynamic objects are known in advance.

\subsection{Environment Representation}
\begin{figure}[t]
\centering
\includegraphics[width=0.63\linewidth]{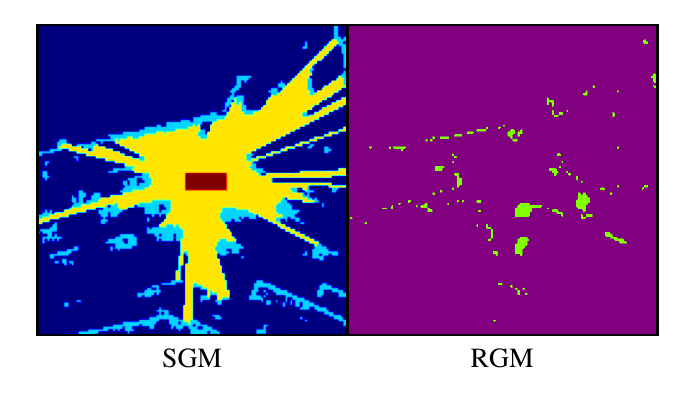}
\vspace{-1em}
\caption{\small An example of SGM (cyan: occupied, yellow: free, blue: occluded, dark red: ego vehicle) and its corresponding RGM (green: change in occupancy states, purple: no change in occupancy states).}
\label{sgm_rgm}
\vspace{-1.5em}
\end{figure}

We use an occupancy-based environment representation for both the static-dynamic object segmentation and the environment prediction modules. For static-dynamic object segmentation, we propose using input in the forms of sensor grid maps (SGMs) and residual grid maps (RGMs), as illustrated in 
\cref{sgm_rgm}.
The OGM representation is used as the input to the environment prediction model, as done by~\citet{mery_icra}. 

The SGMs are occupancy grid maps that reflect sensor measurements, and are generated in the process of converting point clouds to OGMs. Each grid cell contains the occupancy class instead of the belief of its occupancy probability as in the OGM representation. The occupancy classes are occupied, free, or occluded.

We design the RGMs to represent changes in the occupancy class for each grid cell between the current and past time steps.
Our aim is to provide cues to the model to pay closer attention to regions where there are changes in the occupancy class in the same grid cell. The changes are indicative of the presence of potential dynamic objects.

\subsection{Static-Dynamic Object Segmentation}
\label{approach:dynamic_object_segmentation}
Since the nature of the moving object segmentation task is comparable to that of the semantic segmentation task, our segmentation module uses the SalsaNext~\cite{salsanext} architecture, following~\citet{mos}, to learn the environment dynamics in the input data. Unlike their method, our proposed segmentation model is trained on occupancy-based SGMs and RGMs, instead of range image representation, for direct integration with the downstream OGM environment prediction module. The SGMs provide the current occupancy information in the spatial dimension, and the RGMs provide the necessary dynamic information in the temporal dimension. The outputs of the static-dynamic object segmentation module are the discretized dynamic masks, which are used to construct the dynamic and static OGMs as inputs to the downstream environment prediction module.

\subsubsection{SGMs}
SGMs are in $\mathbb{R}^{W \times H}$, where $W$ and $H$ are the width and height. Each cell is assigned either free, occupied, or occluded class. Prior to receiving sensor measurements, the SGMs are initialized to the occluded class for every grid cell to reflect lack of information. Ray tracing is then performed to determine the free space between a sensor measurement and the ego vehicle, and the occupancy class is updated to the free class accordingly. 
Cells corresponding to the sensor measurements are assigned the occupied class. 

\subsubsection{RGMs}
\label{approach:RGMs}
Inspired by~\citet{mos}, who exploit the difference in range images to segment moving objects, we propose using occupancy-based RGMs. The RGMs are created from the current and past SGMs, and are in $\mathbb{R}^{W \times H}$. The past SGMs are transformed into the current local coordinate system to compensate for the ego motion.  
Cells whose occupancy class change from one known class to another (e.g. from free to occupied) are set to $1$, and $0$ otherwise.
We do not consider occluded cells because changes in their occupancy states are not reflective of them being moving object candidates. 
Since the true occupancy of the occluded cells are unknown, they are unsuitable to be used to make decisions about their potential movement.    
The RGM is then concatenated with the current SGM as an extra channel, where the SGM provides spatial information and the RGM encodes temporal information. The output of the model is a discretized binary dynamic mask, $M_{d} \in \mathbb{R}^{W \times H}$, where $1$ corresponds to dynamic objects, and static object otherwise. 

\subsection{Environment Prediction}
\label{approach:env_prediction}
For the spatiotemporal environment prediction task, we use the double-prong model based on the PredNet~\cite{PredNet} architecture, following~\citet{mery_icra}. However, our model uses the dynamic mask outputs from the upstream  module to split the OGM data into static and dynamic OGMs, instead of relying on object detection and tracking information to identify the moving objects. The dynamic OGMs are obtained by multiplying the dynamic masks $M_{d}$ with the OGMs, whereas the static OGMs are obtained by multiplying by $1-M_{d}$. 
The OGM data are generated following the process outlined by~\citet{Masha}.
The model learns both the spatial and temporal structures in the OGMs, and outputs the predicted future OGMs.

Following prior work~\cite{mery_icra}, the local environment around the ego vehicle is discretized into grid cells to generate the evidential OGMs (eOGMs). The eOGM is an alternative representation of OGM created by updating the grid cells using Dempster--Shafer Theory (DST)~\cite{dst}.
Since a grid cell must be either free or occupied, the allowable hypotheses are \emph{free}, \emph{occupied}, or \emph{occluded}, $\{\{F\}, \{O\}, \{O, F\} \}$. Each allowable hypothesis is associated with its corresponding Dempster--Shafer belief mass, which represents the degree of occupancy belief in that cell~\cite{dst}. The belief masses of all allowable hypotheses should sum to unity for each grid cell. The eOGMs are in $\mathbb{R}^{W \times H \times C}$, where $W$, $H$, and $C$ are the width, height, and number of channels. The channels contain the Dempster--Shafer belief masses for the occupied ($m(\{O\}) \in [0, 1]$) and free ($m(\{F\}) \in [0, 1]$) hypotheses. 
Both the static and dynamic eOGMs have $2$ channels, each consisting of its respective $m(\{O\})$ and $m(\{F\})$.

\section{EXPERIMENTS}
\label{experiments}
\subsection{Environment Representation Generation}
The SGMs, RGMs and OGMs are generated from the Waymo Open Dataset~\cite{Sun_2020_CVPR} to have a width and a height of $128$ cells, with a resolution of \SI{0.33}{\metre}, capturing approximately $\SI{42}{\metre} \times \SI{42}{\metre}$ with the ego vehicle fixed at the center. The cell resolution is chosen such that each vehicle is covered by a sufficient number of cells. 
 
\subsection{Model Training}
The inputs to our static-dynamic object segmentation module are the SGMs and RGMs, and the outputs are the dynamic masks. The model is trained to segment the environment into static and dynamic cells given an input of the current SGM and its corresponding RGM. Based on validation set performance, the RGMs generated from SGMs at time $t$, and $t-5$ (\SI{0.5}{\second} earlier), are selected. The train, validation and test split ratios for the dataset are $0.6$, $0.1$, and $0.3$, respectively. We train the network using stochastic gradient descent with an initial learning rate of $0.01$ over $60$ epochs. 

For the environment prediction module, the model receives as input the static and dynamic OGMs, generated from applying the dynamic mask outputs from the static-dynamic object segmentation module to the OGMs. The outputs are the complete OGM predictions of the entire environment surrounding the ego vehicle. The OGMs are arranged into sequences of $20$ consecutive frames each, which represents \SI{2}{\second} of driving data. Following the original experimental setup of~\citet{mery_icra}, the model is trained on an input of the past $5$ OGM frames to predict the future $15$ OGM frames in the same sequence. We train the model in $2$ modes, as suggested by~\citet{PredNet}. The first mode is training the model on the current OGM to predict the OGM at the next time step. The second mode is fine-tuning the model to recursively predict the next $15$ OGMs with the weight parameters initialized from the previous mode of training. The hyperparameters are set following the prior work \cite{mery_icra}. We validate our proposed dynamics-aware environment prediction system by comparing the prediction results with those of the prior work~\cite{mery_icra}.
The inference time is \SI{82}{\milli\second} (\SI{12}{\hertz}) per sample on an Intel i7-5930K at \SI{3.5}{\giga\hertz} and an NVIDIA GeForce TITAN X graphics card.  

\section{Evaluation}
\label{results}
This section summarizes the results of our proposed system evaluated on the Waymo Open Dataset~\cite{Sun_2020_CVPR}. We first present the static-dynamic object segmentation results, followed by the environment prediction results. 

\subsection{Static-Dynamic Object Segmentation}
\label{results:DOS}
\begin{figure}[t]
\centering
\begin{subfigure}[t]{0.47\textwidth}
\centering
\includegraphics[width=1\textwidth]{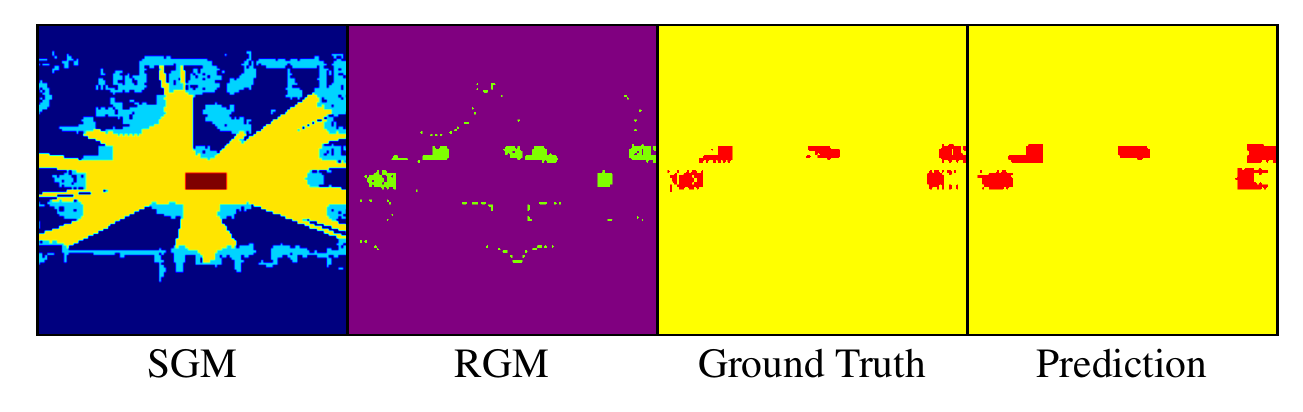}
\vspace{-0.8cm}
\caption{\small Multiple vehicles driving in traffic.}
\label{DOS_a}
\end{subfigure}
\centering
\begin{subfigure}[t]{0.47\textwidth}
\includegraphics[width=1\textwidth]{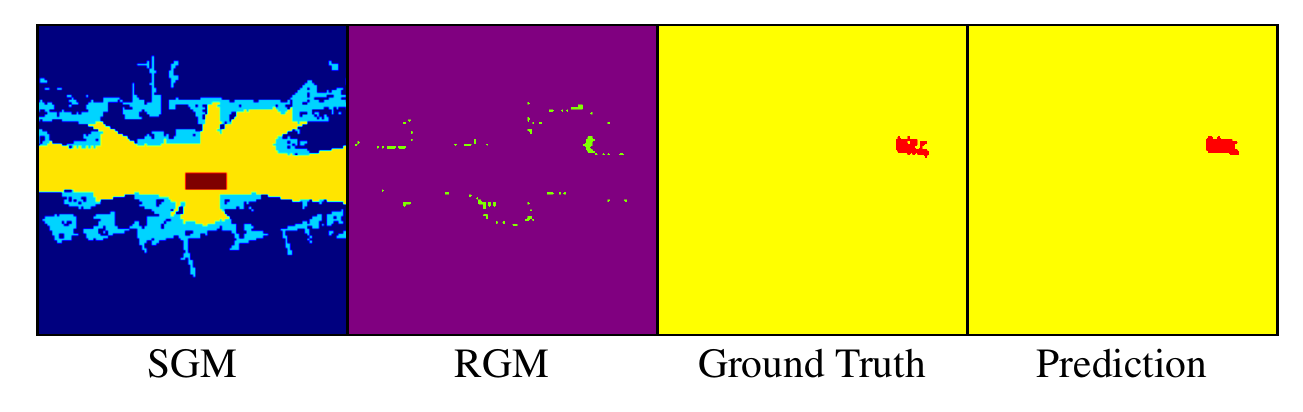}
\vspace{-0.8cm}
\caption{\small An ego vehicle driving through a residential area with multiple parked cars along the side of the street.}
\label{DOS_b}
\end{subfigure}
\begin{subfigure}[t]{0.47\textwidth}
\includegraphics[width=1\textwidth]{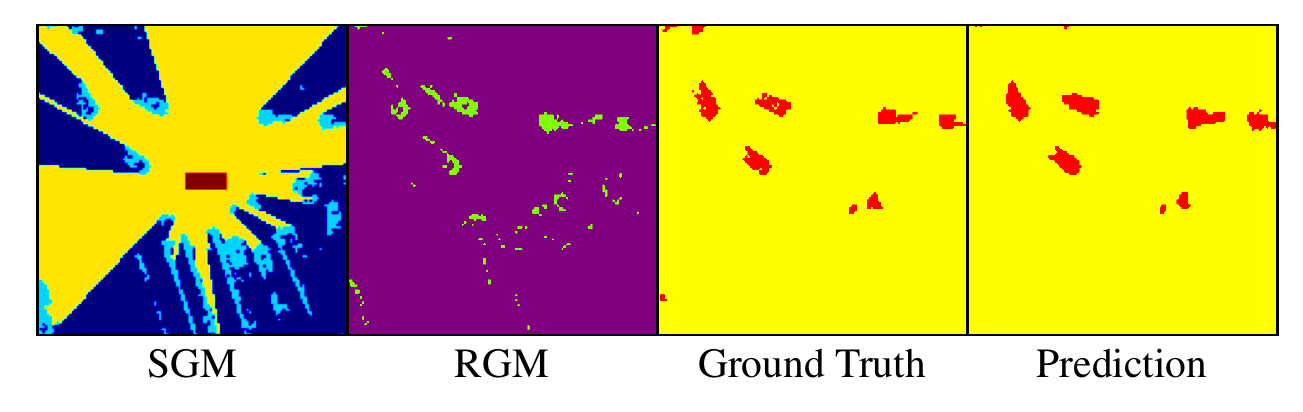}
\vspace{-0.8cm}
\caption{\small An ego vehicle turning at an intersection with pedestrians crossing the street.}
\label{DOS_d}
\end{subfigure}
\begin{subfigure}[t]{0.47\textwidth}
\includegraphics[width=1\textwidth]{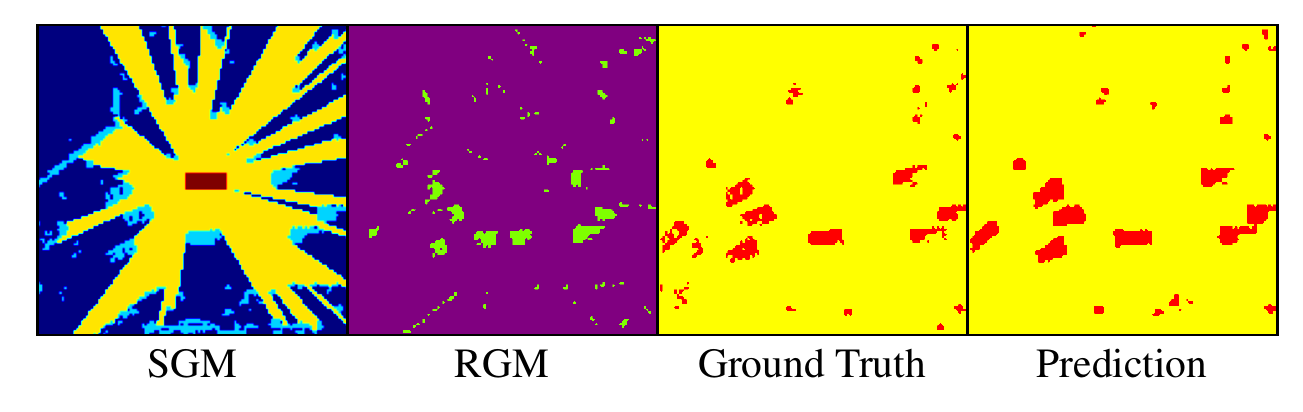}
\vspace{-0.8cm}
\caption{\small An ego vehicle driving through a busy intersection with multiple pedestrians and vehicles.}
\label{DOS_e}
\end{subfigure}
\caption{\small Example dynamic object prediction results. The inputs are SGMs (cyan: occupied, yellow: free, blue: occluded, dark red: ego vehicle) and RGMs (green: change in occupancy states, purple: no change in occupancy states). The prediction outputs (red: dynamic, yellow: static) are shown along with the ground truth labels.}
\label{figure:DOS_qualitative}
\vspace{-2em}
\end{figure}

\begin{figure}[t!]
\centering
\stackinset{l}{2.99in}{b}{2.9in}{\textcolor{white}{\solidcirc[0]{1pt}{0.16}{0.1}}}
{\stackinset{l}{2.99in}{b}{2.285in}{\textcolor{white}{\solidcirc[0]{1pt}{0.16}{0.1}}}
{\stackinset{l}{2.99in}{b}{1.675in}{\textcolor{white}{\solidcirc[0]{1pt}{0.16}{0.1}}}
{\stackinset{l}{2.99in}{b}{1.06in}{\textcolor{white}{\solidcirc[0]{1pt}{0.16}{0.1}}}
{\stackinset{l}{2.99in}{b}{0.45in}{\textcolor{white}{\solidcirc[0]{1pt}{0.16}{0.1}}}
{\stackinset{l}{3.27in}{b}{2.9in}{\textcolor{white}{\solidcirc[0]{1pt}{0.16}{0.1}}}
{\stackinset{l}{3.27in}{b}{2.285in}{\textcolor{white}{\solidcirc[0]{1pt}{0.16}{0.1}}}
{\stackinset{l}{3.27in}{b}{1.675in}{\textcolor{white}{\solidcirc[0]{1pt}{0.16}{0.1}}}
{\stackinset{l}{3.27in}{b}{1.06in}{\textcolor{white}{\solidcirc[0]{1pt}{0.16}{0.1}}}
{\stackinset{l}{3.27in}{b}{0.45in}{\textcolor{white}{\solidcirc[0]{1pt}{0.16}{0.1}}}
{\input{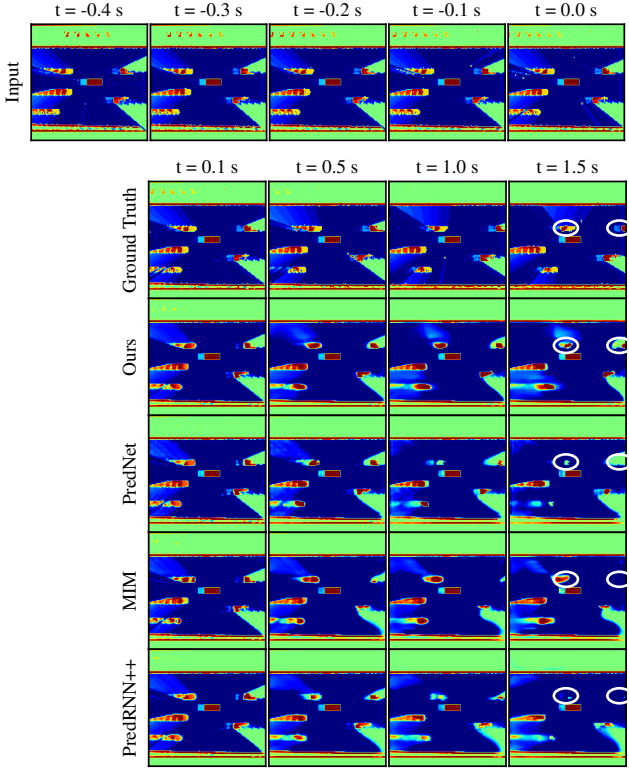}}}}}}}}}}}
\caption{\small Example OGM predictions (red: occupied, green: occluded, blue: free) of a scene with multiple vehicles. The predictions are shown at \SI{0.1}{\second}, \SI{0.5}{\second}, \SI{1.0}{\second}, and \SI{1.5}{\second} ahead.} 
\vspace{-1.5em}
\label{fig:env_pred}
\end{figure}

\begin{table*}[t!]
    \centering
    \caption{\small The averaged MSE, dynamic MSE, and IS metrics on OGM predictions over the entire prediction horizon. Lower is better.}
    \begin{tabular}{@{}lrrrr@{}}
        \toprule
        \multicolumn{1}{c}{\bf Models} &
        \multicolumn{1}{c}{\bf Parameters $\times 10^{6}$} &
        \multicolumn{1}{c}{\bf MSE $\times 10^{-2}$} &
        \multicolumn{1}{c}{\bf Dynamic MSE$\times 10^{-3}$} &
        \multicolumn{1}{c}{\bf IS}\\
        \midrule
        MIM~\cite{mim} & $6.8$ & $3.41 \pm 0.0007$ &  $\mathbf{2.23\pm 0.0002}$ & $6.86 \pm 0.066$  \\
        PredRNN++~\cite{predrnn++} & $1.8$ & $3.45 \pm 0.0008$ & $2.38\pm 0.0002$ & $7.08 \pm 0.070$\\
        PredNet~\cite{PredNet} & $1.2$ & $3.50\pm 0.0008$  & $2.53\pm 0.0003$ & $7.58 \pm 0.074$\\
        \midrule
        Ours (Double-Prong + Predicted Dynamic Masks) & $6.7 + 1.8$ & $\mathbf{3.19 \pm 0.0007}$ & $2.24 \pm 0.0002$ & $\mathbf{5.97 \pm 0.057}$\\
        Double-Prong + Ground Truth Dynamic Masks~\cite{mery_icra} & $1.8$ & $\mathbf{3.17 \pm 0.0007}$ & $\mathbf{2.03 \pm 0.0002}$ & $\mathbf{5.86 \pm 0.058}$\\
        \bottomrule
    \end{tabular}
    \label{table:metrics}
\vspace{-2em}
\end{table*}

\cref{figure:DOS_qualitative} shows example dynamic object segmentation results using the SalsaNext~\cite{salsanext} semantic segmentation model with the proposed occupancy-based SGMs and RGMs as an input representation. The ground truth labels show the target dynamic predictions. Occupied, free, and occluded cells in the SGMs are depicted in cyan, yellow, and blue, respectively. The ego vehicle, depicted in dark red, is fixed at the center, and is moving to the right in the SGMs. Cells with changes between the known occupancy classes (free or occupied) are shown in green in the RGMs, and purple otherwise. We only consider cells with known occupancy classes to be valid for computing the RGMs. In the ground truth labels and dynamic mask prediction outputs, moving and static cells are depicted in red and yellow, respectively.

As illustrated in \cref{DOS_a,DOS_b,DOS_d}, our model is able to segment all of the moving vehicles present, and accurately capture their positions and orientations for a variety of traffic scenes. Similarly, static objects, including stopped vehicles in traffic, parked vehicles on the side of the road, and buildings, are correctly classified as static. We observe that the moving vehicles have very strong signals (represented as larger number of cells with changes in its occupancy class, depicted in green) in the RGM, which indicates to the network that these cells likely correspond to the dynamic cells. Static objects, on the other hand, accordingly have weaker signals. We hypothesize that the clear differences in the signal strengths help guide the network in the predictions. 

We also note that, for slower-moving vehicles, such as the one shown in the residential area scene in \cref{DOS_b}, the signal strength is much weaker. However, our network is able to accurately classify the vehicle as dynamic, among all other parked vehicles on the side of the street (shown in the SGM). We hypothesize that similar examples exist in the training data, enabling the network to learn and recognize between weaker signals belonging to static objects and slow vehicles.

Classifying very slow-moving and small objects, like pedestrians, can be particularly challenging. Our model is able to segment the majority of the walking pedestrians, shown as a cluster of smaller-sized moving cells in \cref{DOS_d} and \cref{DOS_e}, as dynamic. These success cases occur when there are signals in the RGM inputs, even when the signals are relatively weaker compared to moving vehicles. On the other hand, our model misses the prediction when the signal is absent. This behavior is observed in the moving vehicles as well. In \cref{DOS_e}, the dynamic prediction misses two pedestrians (top right), and two vehicles (bottom left). These objects show no signal in the RGM, and we hypothesize that this could be a result of very slow motion, or occlusion. 

\subsection{Occupancy Grid Prediction}
\subsubsection{Qualitative Results}
\cref{fig:env_pred} illustrates example predictions using our proposed dynamics-aware OGM prediction approach. Our predictions are compared against the PredNet~\cite{PredNet}, MIM~\cite{mim}, and PredRNN++~\cite{predrnn++} baselines. The inputs to the model are the past $5$~OGMs (depicted on the top row  in \cref{fig:env_pred}), and the predicted static-dynamic masks obtained from the upstream static-dynamic object segmentation module. Occupied, occluded, and free cells are shown in red, green, and blue, respectively. The ground truth OGM labels are shown at selected prediction times. The ego vehicle is fixed at the center and is moving to the right. 

\cref{fig:env_pred} shows an example scene with five vehicles, excluding the ego vehicle, traveling at relatively high speeds, as suggested by the long trails. Our model retains all the vehicles throughout the \SI{1.5}{\second} prediction horizon, and accurately capture their motions. Due to an accumulation of prediction errors, the predictions become less accurate with time.

\subsubsection{Quantitative Results}
To quantitatively evaluate the performance of our dynamics-aware environment prediction module, we use the mean squared error (MSE) and the image similarity (IS)~\cite{im_sim} metrics between the predicted and target OGMs. The MSE metric is used to assess how well the predicted occupancy probability for each cell corresponds to its ground truth value.
The IS metric is used to measure how well the structure of the scene is maintained in the OGM predictions.
To calculate the IS metric, the minimum Manhattan distance is calculated between two grid cells (one from the target OGM and the other from the predicted OGM) with the same occupancy classes (occupied, free, and unknown). 
The dynamic MSE~\cite{mery_icra} is also computed to assess the performance on predicting the future occupancy probability of only moving object cells. To calculate the dynamic MSE, the dynamic masks are applied to the target and predicted OGMs, then we calculate the MSE.

\cref{table:metrics} shows the averaged MSE, dynamic MSE, and IS metrics computed over the entire $15$ prediction time steps. We benchmark our proposed method against three other models, namely MIM~\cite{mim}, PredNet~\cite{PredNet}, and PredRNN++~\cite{predrnn++}, as well as compare with the original double-prong environment prediction model~\cite{mery_icra}.
The additional $6.7 \times 10^{6}$ number of parameters in our method comes from the static-dynamic segmentation model.

Our model performs better than the baseline models across the MSE and IS metrics, and is comparable to the MIM~\cite{mim} network in the dynamic MSE metric ($0.4$\% difference).
Our method outperforms the MIM~\cite{mim}, PredRNN++~\cite{predrnn++}, and PredNet~\cite{PredNet} baseline methods in MSE by about $6.5$\%, $7.5$\%, and $8.9$\%, respectively. 

The performance of the original double-prong model~\cite{mery_icra} is an upper bound to the performance that our model can achieve, since they use the ground truth static-dynamic object segmentation, while we use the predicted dynamic masks. Our model performance is comparable to the original double-prong model in the MSE ($0.62$\% difference) and IS metrics ($1.8$\% difference). Despite not being as close to the upper bound in the dynamic MSE, we still perform relatively better compared to the baseline models. 

We posit that using predicted dynamic masks, instead of the ground truth dynamic masks, has a direct effect on predicting the OGMs in the dynamic cells, compared to static cells. 
Since the MSE and IS metrics are comparable to the prior double-prong work~\cite{mery_icra}, our upstream static-dynamic object segmentation model accurately segments static and dynamic objects, resulting in the MSE and IS metrics to be very close to the upper bound values.

However, for challenging dynamic objects (e.g. pedestrians, very slow-moving vehicles and occluded objects), our static-dynamic segmentation module could have missed these objects in the segmentation completely, or partially detect them. The partial dynamic object segmentation results in missing some moving cells when assigning them to the dynamic state in the dynamic masks, or not being able to predict their locations accurately. This is corroborated when we calculate the intersection over union (IoU)~\cite{Everingham2009ThePV} metric on the predicted dynamic masks.
The IoU is used to measure how well the predicted dynamic cells overlap with the corresponding ground truths.  
Our IoU metric over the static and moving objects are $99.5$\% and $54.5$\%, respectively, and the average IoU is $77.0$\%. Hence, the difference in the exact location of dynamic cells in the ground truth and predicted static-dynamic object segmentation plays a role when computing the dynamic MSE on the future OGM prediction on dynamic cells, by applying the dynamic masks. However, this discrepancy does not significantly affect the overall OGM prediction accuracy (MSE) and the ability to maintain the structure of the scene in the predictions (IS).

\section{CONCLUSION}
\label{conclusion}
We present a dynamics-aware spatiotemporal occupancy prediction method for future local environment prediction around the ego vehicle. Our approach combines static-dynamic object segmentation and environment prediction together into one system, without resorting to access to highly accurate object detection and tracking information. We show that using an occupancy-based environment representation, along with our proposed residual grid maps, enables accurate static-dynamic object segmentation and offers direct integration with downstream environment prediction. Our model has comparable performance to the upper bound double-prong model that uses ground truth dynamics information, showing that we are able to successfully integrate static-dynamic object segmentation and environment prediction capabilities. 
Future work will consider incorporating semantic segmentation as well. We hypothesize that the model can perform better if given the ability to learn the semantics in the scene, which can help with predicting the motion models of different object types.  

\section*{ACKNOWLEDGMENTS}
We thank Masha Itkina and Ransalu Senanayake for fruitful discussions, and early support and guidance.  
\renewcommand*{\bibfont}{\small}
\printbibliography

\end{document}